\documentclass{article} % For LaTeX2e
\usepackage{iclr2017_conference,times}
\usepackage{hyperref}
\usepackage{afterpage}
\usepackage{url}
\usepackage{csvsimple}

\usepackage{svg}
\usepackage{natbib}
\usepackage{graphicx}

\usepackage{color}

\newcommand{\cut}[1]{}

\newcommand{\note}[1]{}
\renewcommand{\note}[1]{~\\\frame{\begin{minipage}[c]{\textwidth}\vspace{2pt}\center{#1}\vspace{2pt}\end{minipage}}\vspace{3pt}\\}

\title{A Downsampled Variant
of ImageNet as an Alternative to the CIFAR datasets}

\author{Patryk Chrabaszcz, Ilya Loshchilov \& Frank Hutter \\
University of Freiburg\\
Freiburg, Germany,\\
\{chrabasp,ilya,fh\}@cs.uni-freiburg.de\\
}

\begin{document}
% TODO: Change naming from Imagenet32 to Imagenet32x32 etc. 
% Mention that we have 8x8 and 128x128
% Include experiments with 8x8 ?? 

\maketitle

\begin{abstract}
The  original ImageNet dataset is a popular large-scale benchmark for training Deep Neural Networks. Since the cost of performing experiments (e.g, algorithm design, architecture search, and hyperparameter tuning) on the original dataset might be prohibitive, we propose to consider a downsampled version of ImageNet. In contrast to the CIFAR datasets and earlier downsampled versions of ImageNet, our proposed ImageNet32x32 (and its variants ImageNet64x64 and ImageNet16x16) contains exactly the same number of classes and images as ImageNet, with the only difference that the images are downsampled to 32$\times$32 pixels per image (64$\times$64 and 16$\times$16 pixels for the variants, respectively). Experiments on these downsampled variants are dramatically faster than on the original ImageNet and the characteristics of the downsampled datasets with respect to optimal hyperparameters appear to remain similar. The proposed datasets and scripts to reproduce our results are available at %\linebreak
\linebreak \url{http://image-net.org/download-images} and $\;\;\;\;$ \linebreak   \url{https://github.com/PatrykChrabaszcz/Imagenet32_Scripts}
\end{abstract}

\section{Introduction}

Deep learning research has been substantially facilitated by the availability of realistic and accessible benchmark datasets, such as CIFAR-10 and CIFAR-100 \citep{krizhevsky2009learning} (and MNIST~\citep{lecun1998gradient} in the 1990s). With the progress of machine learning, simple datasets lose some of their relevance, and more complex datasets/tasks become more important. While good results can be achieved on more complex datasets, such as ImageNet~\citep{krizhevsky2012imagenet,ILSVRC15}, this incurs a large computational burden, making it intractable to achieve state-of-the-art performance without massive compute resources
(training a strong ImageNet model typically requires several GPU months). 

Due to this computational expense of running experiments on the original ImageNet dataset we propose to explore cheaper alternatives that preserve the dataset's complexity. 
In order to check the scalability of new methods, neural architectures and hyperparameters associated with them, one might be interested in a downscaled version of ImageNet which allows for cheaper experimentation. Moreover, a lower resolution of the images would make the classification task much more difficult and would thus postpone the saturation of benchmarking results currently observed on CIFAR-10, e.g., 3\% error obtained by ~\cite{shakeshake2017} compared to roughly 6\% obtained by a trained human \citep{KarpathyCIFAR10}. 

To address this issue, we provide downsampled variants of the original ImageNet dataset and analyze results on them w.r.t.\ different hyperparameter settings and network sizes. We obtain surprisingly strong classification results on our downsampled variants and find qualitative results to be very similar across downsampling sizes. This suggests that these downsampled datasets are useful for facilitating cheap experimentation.

The basic contributions of this report are as follows:
\begin{itemize}
\item We make available downsampled versions of ImageNet ($64\times 64$, $32\times 32$, and $16\times16$ pixels) to facilitate fast experimentation with different network architectures, training algorithms, and hyperparameters.
\item We show that different downsampling techniques yield similar results, except for a nearest neighbor approach, which performed worse in all our experiments.
\item Using Wide ResNets~\citep{zagoruyko2016wide}, we obtain surprisingly good performance, matching the baseline by the pioneering AlexNet~\citep{krizhevsky2012imagenet} (18.2\% top-5 error) while using ImageNet32x32 (whose images have roughly 50$\times$ less pixels per image than the original ones). 
\item We show that the range of optimal learning rates does not change much across ImageNet16x16, ImageNet32x32, and ImageNet64x64, as well as across different network widths. This could be exploited by multi-fidelity methods for architecture and hyperparameter search~\citep{li2016hyperband,klein2016fast}.
\end{itemize}

\section{Downsampling ImageNet}\label{sec:downsampling}

The original ImageNet dataset consists of images released as a part of the ILSVRC-2012 classification dataset \citep{krizhevsky2012imagenet,ILSVRC15}. Each image belongs to one of 1000 object classes, with the number of training images per class varying from 732 to 1300; there are 50 validation images per class. The size of the original images varies; therefore, a preprocessing step is usually applied to scale and crop images to the size of 224 $\times$ 224 pixels.

We are aware of two datasets that contain low resolution images derived from the ImageNet dataset: 
\begin{itemize}
\item \textbf{Downsampled ImageNet}~\citep{oord2016pixel}, like our datasets, contains all images in ImageNet, but since it was constructed for unsupervised learning, it does not provide the actual image labels and can therefore not be used for supervised learning.
\item \textbf{TinyImageNet} (available at \url{https://tiny-imagenet.herokuapp.com/})
contains a subset of 200 classes with 500 images per class.
\end{itemize}

\cite{mishkin2016systematic} suggested to use 128x128 pixels  ImageNet images to evaluate various deep learning  techniques, but their dataset is not available. 

We downsample / resize the original images to smaller images of 32x32 pixels to form ImageNet32x32, to images of 64x64 pixels to form ImageNet64x64 and to images of 16x16 pixels to form ImageNet16x16. In contrast to TinyImageNet, we do not reduce the number of classes and number of images. All images are shuffled and then divided into 10 different files so that each file is expected to have images from all classes. The validation data is stored in a separate file, both the training and validation data points are labeled (e.g., indexed starting from 1) according to the mapping file of the  ImageNet devkit. Each file contains images, labels and the mean image computed over the whole training set. We keep the same format of files as the one that is commonly used for the CIFAR datasets. ImageNet16x16, ImageNet32x32 and ImageNet64x64 take 1 GB, 4 GB  and 16 GB of disk space, respectively.

\begin{figure}[b]
\begin{center}
\includegraphics[width=0.7\textwidth]{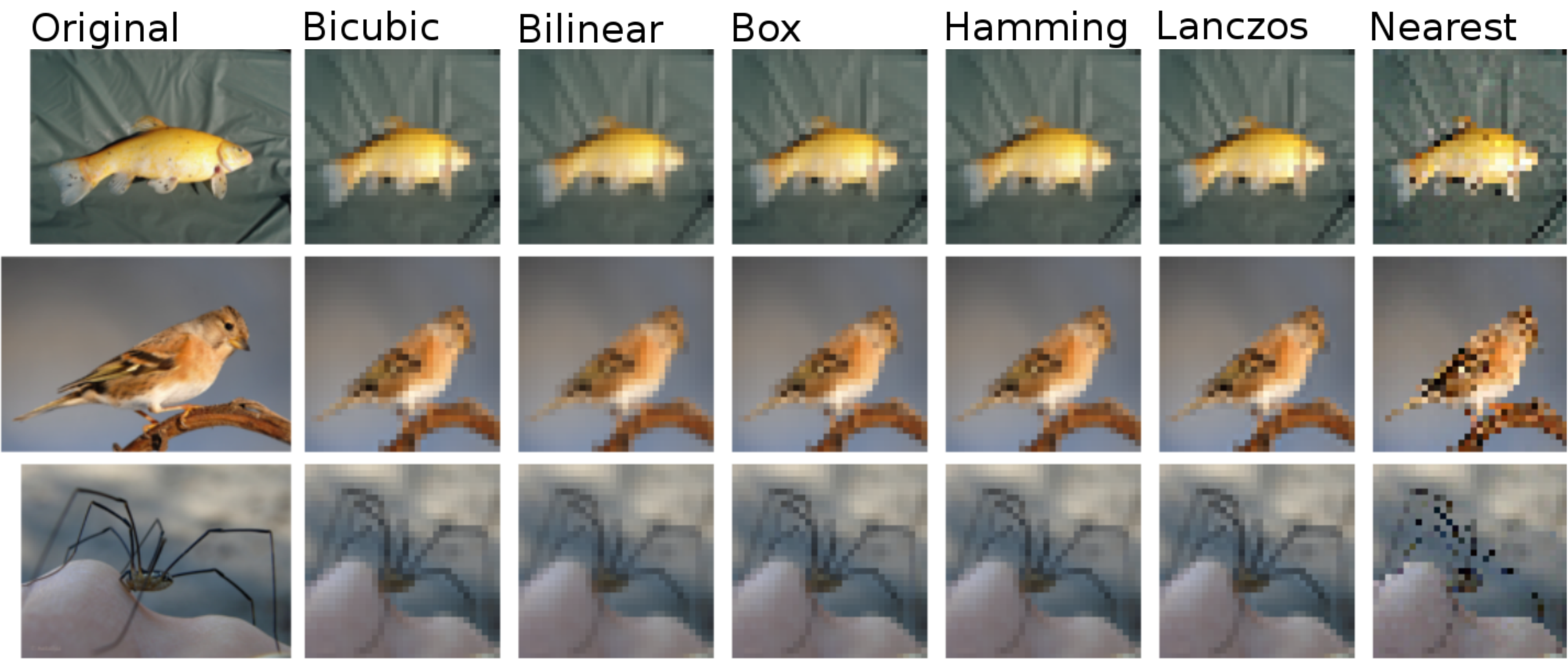}
\end{center}
\caption{The original images (first column) and images obtained by 6 downsampling techniques (left to right): bicubic, bilinear, box, hamming, lanczos, nearest. Our resizing procedure changes the aspect ratio of images.}
\label{Figure1}
\end{figure}

We consider 6 different downsampling techniques available in the \textit{Pillow} library\footnote{Pillow version 4.1 available at \url{https://python-pillow.org}}: lanczos, nearest, bilinear, bicubic, hamming, box (see Figure \ref{Figure1}). In order to check the quality of the downsampled images we use them to train Wide Residual Networks (WRNs) by \cite{zagoruyko2016wide}, expecting that better validation errors will tend to be achieved with downsampling techniques that lose less information.

\section{Experimental setup}

%\textbf{Experimental setup.} 
We train Wide Residual Networks WRN-N-k by \cite{zagoruyko2016wide}, where $N$ is the number of layers and $k$ is a multiplicative factor for the number of filters, with $k=1$ corresponding to 16 filters in the first residual block; increasing $k$ makes the network wider. We use Stochastic Gradient Descent with momentum factor 0.9, drop the learning rate by a factor of 5.0 every 10 epochs, and train up to a total budget of 40 epochs. Throughout, we show validation error rates obtained after training for 31 epochs (right after the last drop of the learning rate). 

Our experiments on ImageNet32x32 employ the original WRNs designed for the CIFAR datasets with 32 $\times$ 32 pixels per image. To adapt WRNs for images with 64 $\times$ 64 pixels per image as used in ImageNet64x64, we add an additional stack of residual blocks to reduce the spatial resolution of the last feature map from 16 $\times$ 16 to 8 $\times$ 8 and thus double the number of features. Analogously, for ImageNet16x16, we remove the last stack of residual blocks. For data augmentation, we  flip images horizontally and concatenate them to the original images, effectively doubling the number of images per epoch. We also use random image shifts (up to 4 pixels horizontally and vertically).

\section{Results}
% on influence of hyperparameter settings and network size

\begin{figure}[tbp]
\begin{center}
%\framebox[4.0in]{$\;$} 
\includegraphics[width=0.49\textwidth]{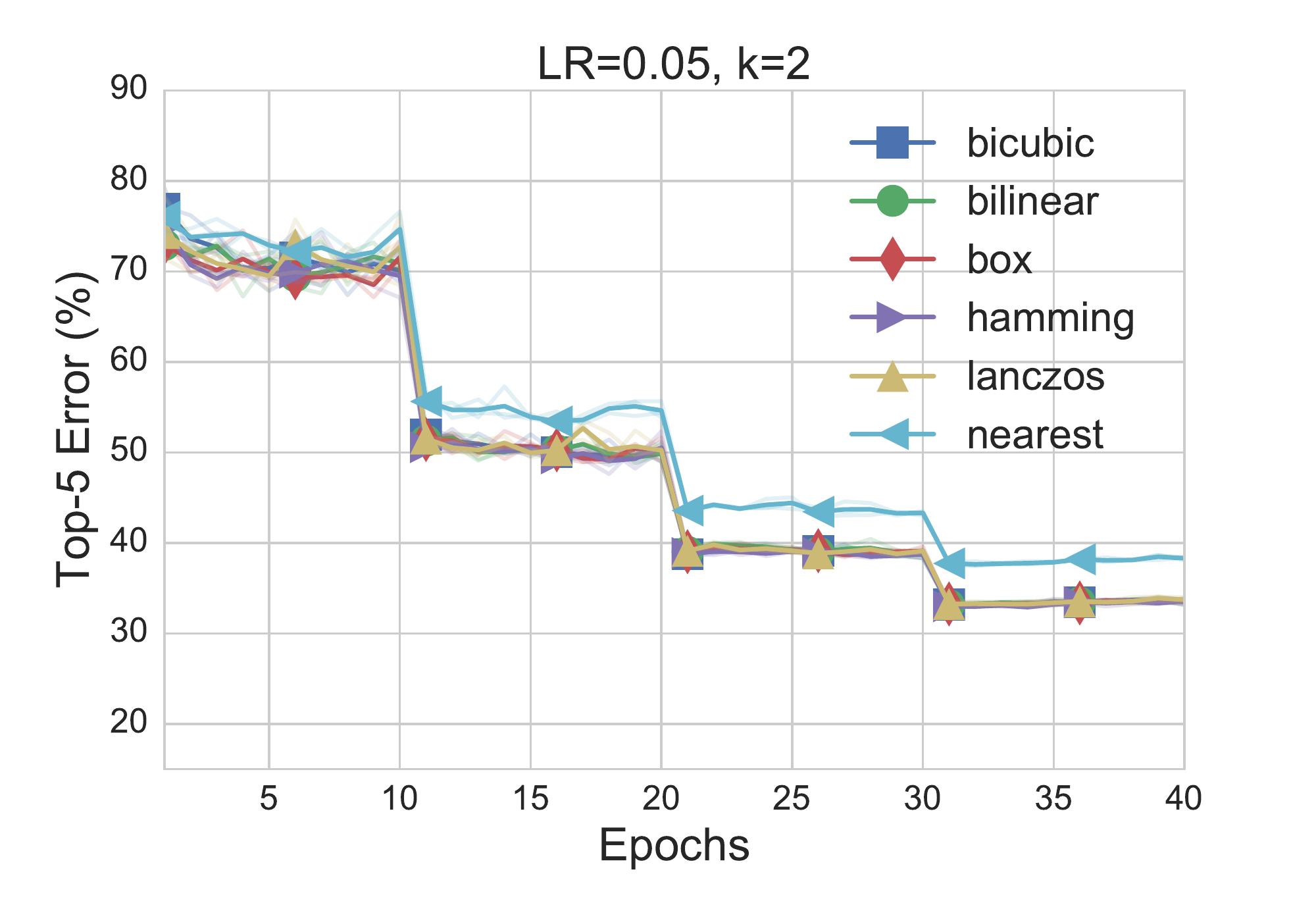}
\includegraphics[width=0.49\textwidth]{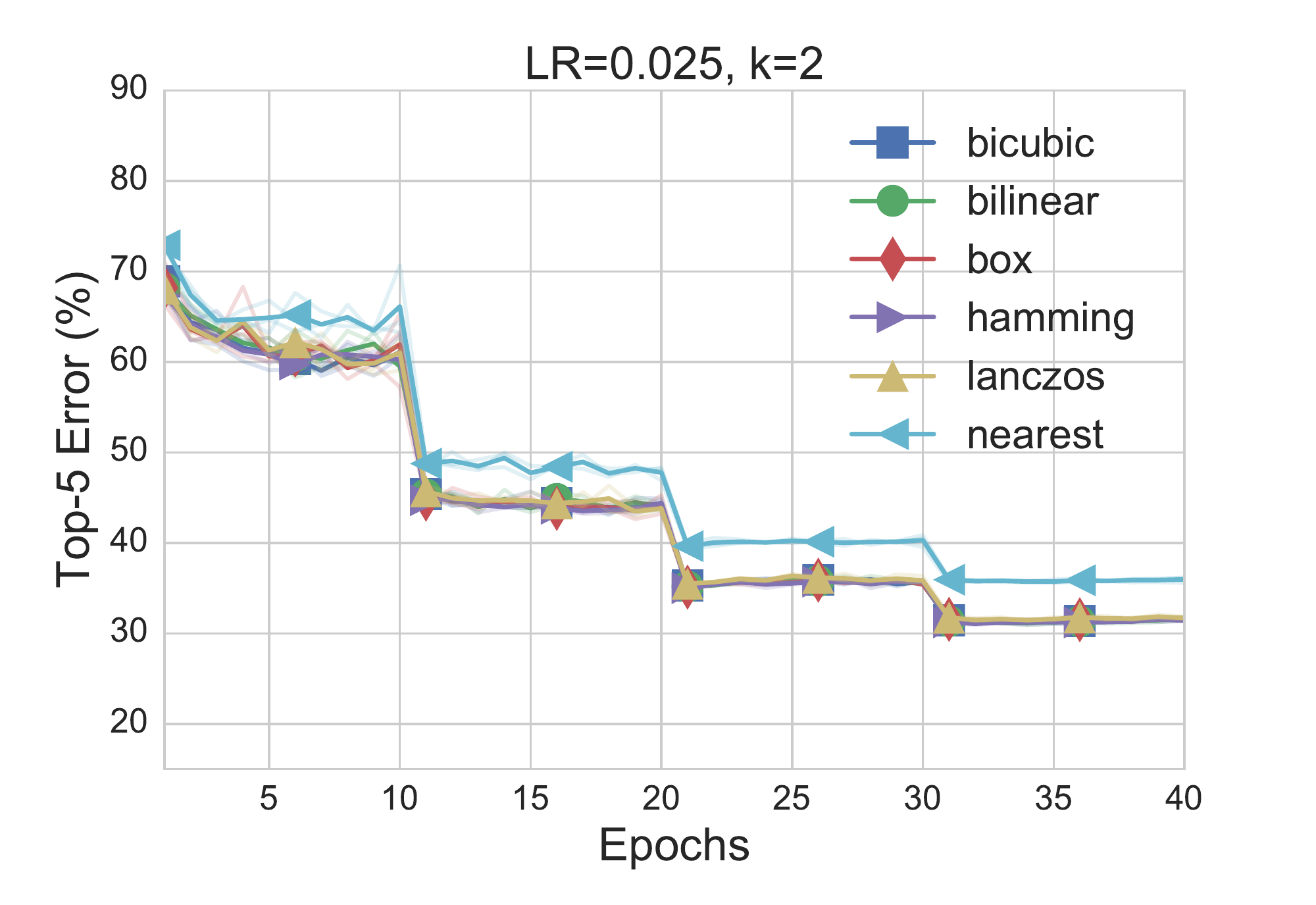}
\includegraphics[width=0.49\textwidth]{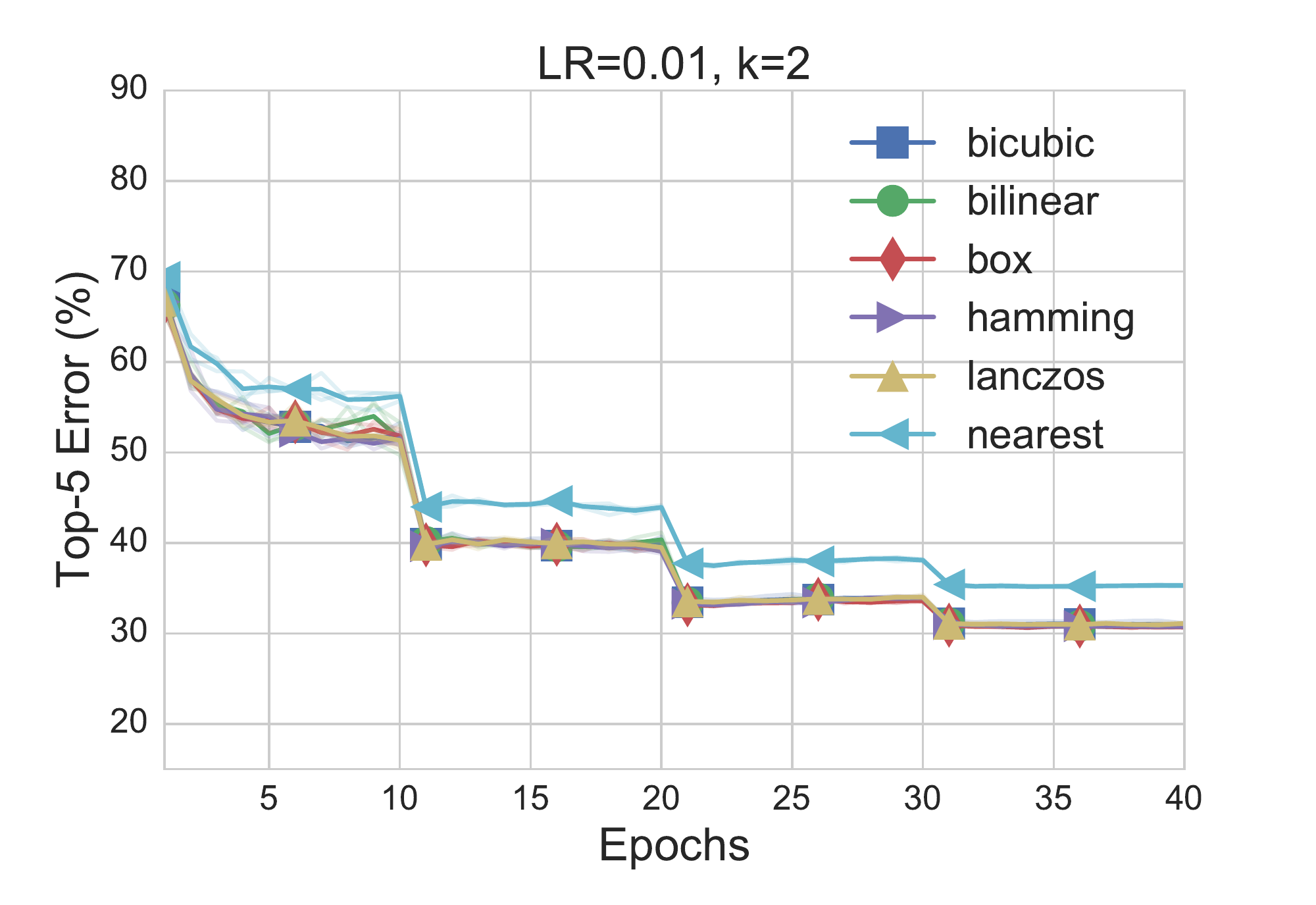}
\includegraphics[width=0.49\textwidth]{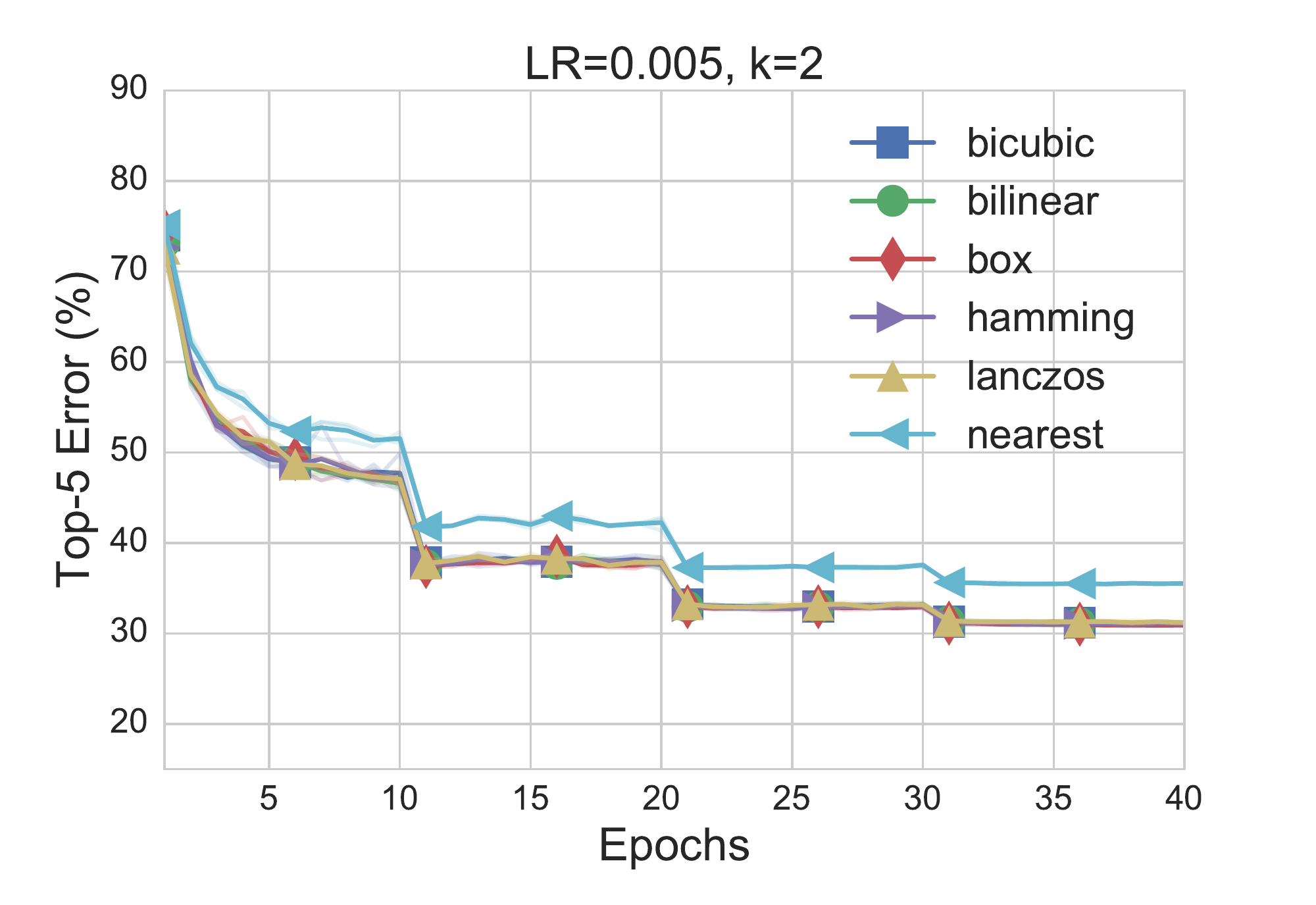}
\includegraphics[width=0.49\textwidth]{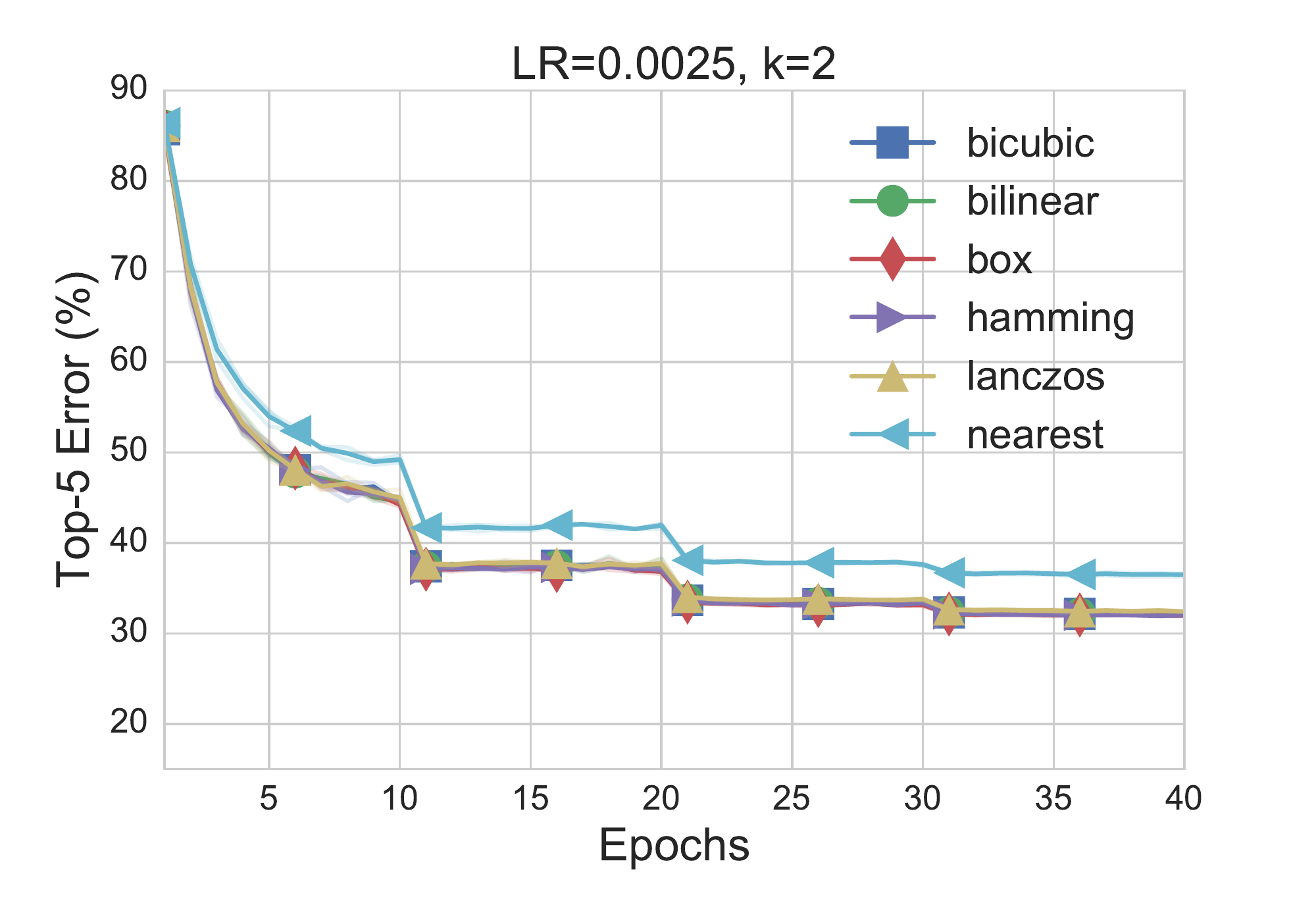}
\includegraphics[width=0.49\textwidth]{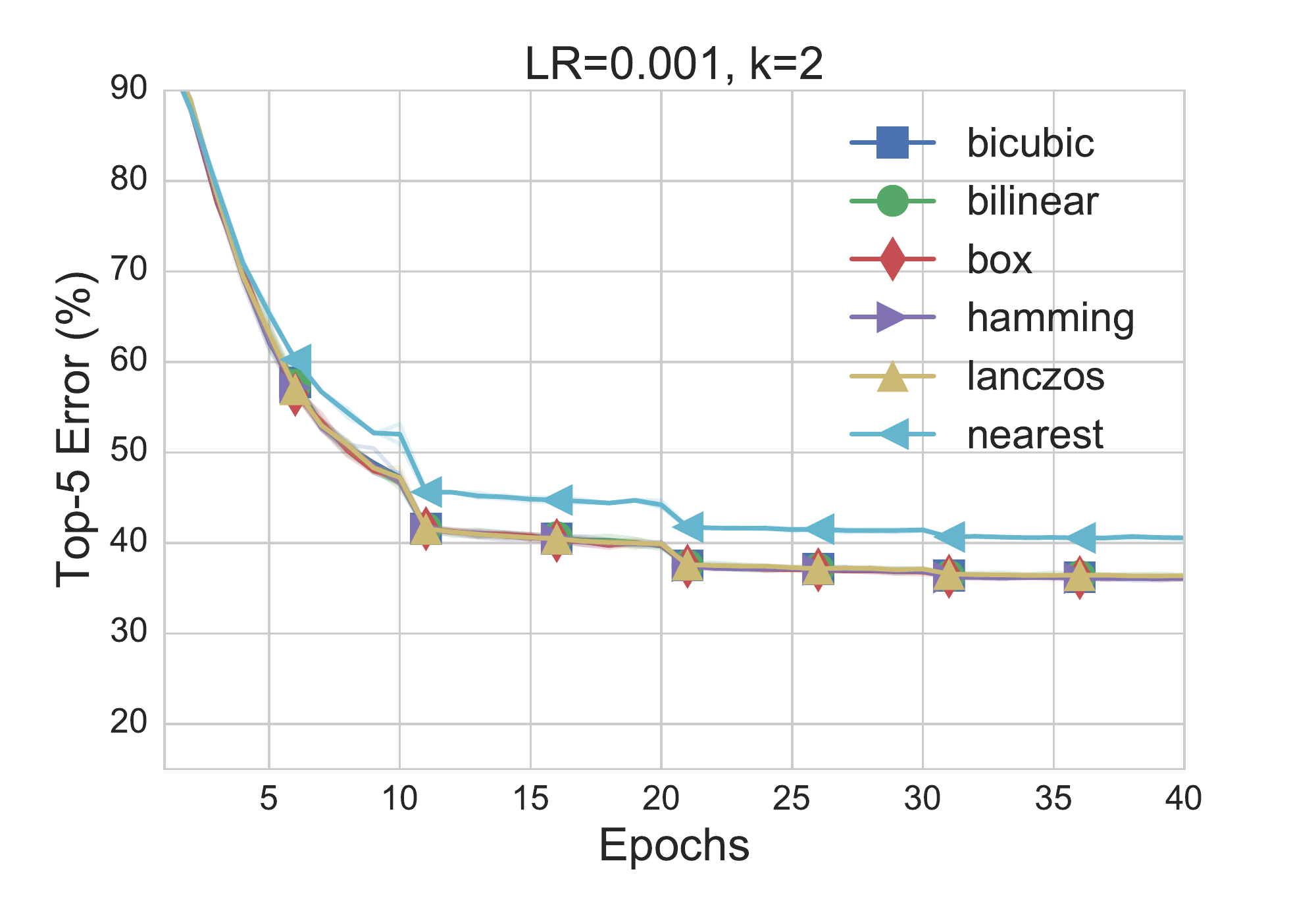}
\end{center}
\caption{The mean Top-5 errors obtained in 3 runs by WRN-28-2 on ImageNet32x32 for different learning rates (indicated at the top of each subfigure as LR) and downsampling algorithms.}
\label{Figure2}
\end{figure}

\textbf{Does the downsampling technique matter?} 
We evaluated the six downsampling techniques described in Section \ref{sec:downsampling} using a small WRN-28-2 network and various initial learning rates $LR \in \{0.001, 0.0025, 0.005, 0.01, 0.025, 0.05\}$. The results in Figure \ref{Figure2} show that all downsampling techniques performed very similarly, except for the nearest neighbour technique which yielded the worst results for all learning rates. This observation is in line with Figure \ref{Figure1} and also holds for ImageNet16x16 and ImageNet64x64 (results not shown for brevity). For all remaining experiments in this paper, we used the box method.

\textbf{Do conclusions drawn for cheap evaluations carry over to expensive ones?} 
Next, we studied to which extent conclusions drawn for small networks and downsampled images carry over to larger networks and higher resolution images. This in turn determines the usefulness of these techniques for speeding up the experimental loop of architecture design and hyperparameter optimization. For this, we performed three experiments:

\begin{itemize}
	\item We studied how the results scale with the network size, more specifically, network width, defined by $k$. Table \ref{Table1} shows that larger $k$ yielded better results independently of the downsampling size.
    %(of course at the price of substantially higher computational expense). 
    Performance on our downsampled datasets was surprisingly strong; for example, on ImageNet32x32, using $k=10$ achieved 40.96\% Top-1 validation error and 18.87\% Top-5 validation error. Interestingly, this matches the original results by AlexNets~\citep{krizhevsky2012imagenet} (40.7\% and 18.2\%, respectively) on full-sized ImageNet (which has roughly 50 times more pixels per image). Clearly, greater image resolution yielded better results (e.g., 12.64\% top-5 performance for ImageNet64x64).

\begin{table}[tbp]
\renewcommand{\arraystretch}{1.2}
\begin{center}
  \begin{tabular}{| l | c | c | c | c | c |}
    \hline
		& width $k$ & \# params & Top-1 error & Top-5 error & Time [days] \\ \hline
		WRN-20-k on ImageNet16x16	& 1 &  0.12M &  85,18\% &  66,12\% & 0.2\\
		WRN-20-k on ImageNet16x16	& 2 &  0.42M &  77,00\% &  54,22\% & 0.4\\
		WRN-20-k on ImageNet16x16	& 5 &  2.3M &  66,60\% &  41,59\% & 1.0\\
		WRN-20-k on ImageNet16x16	& 10 &  8.9M &  59.94\% &  35.10\% & 2.7\\ \hline
		WRN-28-k on ImageNet32x32	& 0.5 &  0.13M &  79,83\% &  57,64\%  & 0.5\\
		WRN-28-k on ImageNet32x32	& 1 &  0.44M &  67,97\% &  42,49\%  & 0.8\\
        WRN-28-k on ImageNet32x32	& 2 &  1.6M &  56,92\% &  30,92\%  & 1.5\\
        WRN-28-k on ImageNet32x32	& 5 &  9.5M &  45,36\% &  21,36\%  & 4.9\\
        WRN-28-k on ImageNet32x32	& 10 &  37.1M &  40,96\% &  18,87\% & 13.8\\ \hline
        WRN-36-k on ImageNet64x64	& 0.5 &  0.44M &  62,35\%  &  36,06\% & 2.1\\
        WRN-36-k on ImageNet64x64	& 1 &  1.6M &  49,79\%  &  24,17\% &  3.4\\
        WRN-36-k on ImageNet64x64	& 2 &  6.2M &  39,55\% &  16,57\%  & 6.4\\
        WRN-36-k on ImageNet64x64	& 5 &  37.6M &  32,34\% &  12,64\%  & 22\\
    \hline
  \end{tabular}
\end{center}
\caption{The mean Top-1 and Top-5 test error rates obtained in 3 runs by WRNs measured right after the last drop of the learning rate, i.e., after epoch 31 (for bigger models training for more than one epoch after the last drop can lead to overfitting). All results are based on a learning rate of 0.01. The timing results are reported for training on a single Titan X GPU.}
\label{Table1}
\end{table}

\item We studied how optimal learning rates changed across different combinations of downsampling sizes and network widths. Figure \ref{Figure3} shows that the region of optimal learning rates remained similar across all our experimental setups, including networks whose space and time complexity differed by up to a factor of 100.
%(ImageNet16x16, $k=1$ vs\ ImageNet32x32, $k=10$). 
%(ImageNet16x16, $k=1$ vs\ ImageNet64x64, $k=5$). 
Additionally, Figure \ref{fig:three_d_plots} compares performance as a function of both learning rate and width multiplier $k$ for downsampling sizes of 32x32 and 16x16, showing qualitatively very similar results in both cases, including the interaction effect that larger values of $k$ favor somewhat larger learning rates than smaller $k$. 
This suggests that small networks and 
downsampled images may indeed facilitate faster experimentation. 

\begin{figure}[tbp]
\begin{center}
\includegraphics[width=0.45\textwidth]{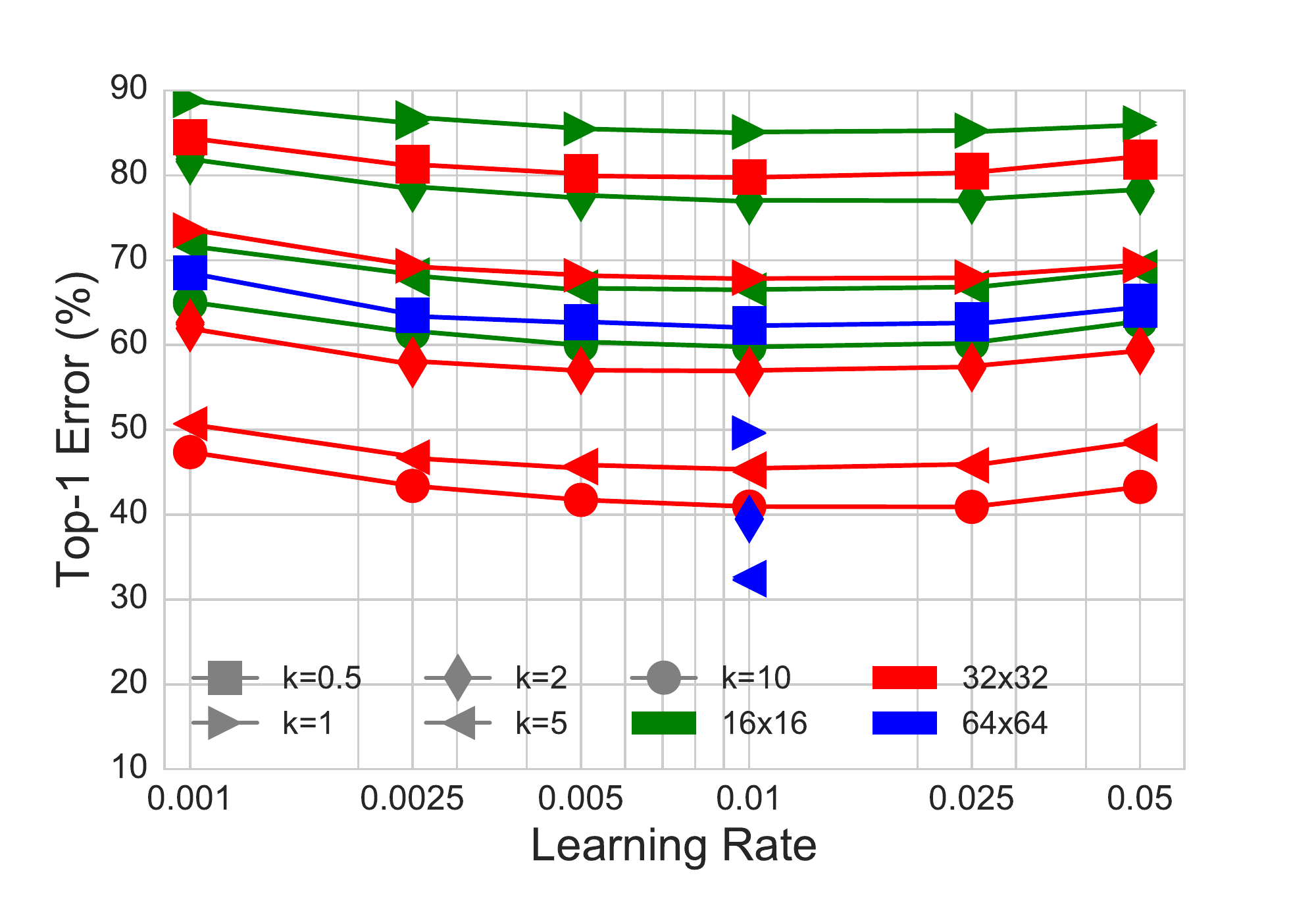}
\includegraphics[width=0.45\textwidth]{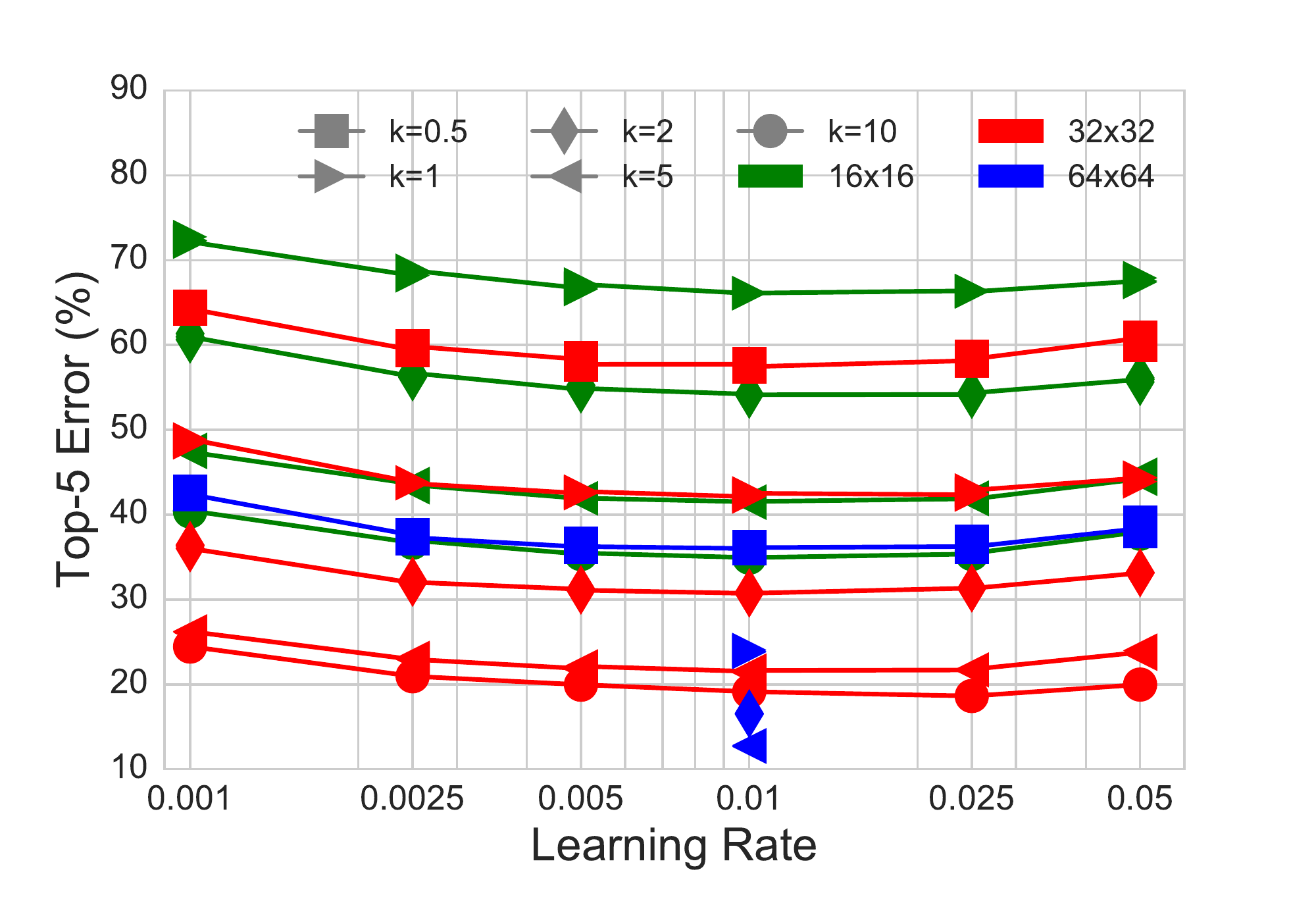}
\end{center}
\caption{The mean Top-1 (Left) and Top-5 (Right) errors obtained in 3 runs by WRN-N-$k$ after 31 epochs with different settings of the initial learning rates and different sizes of the downsampled images (ImageNet16x16, ImageNet32x32 and ImageNet64x64). The results for ImageNet64x64 are shown for different $k$ but a single value of the initial learning rate LR=0.01 which seems reasonably good across different settings.\label{Figure3}}
\end{figure}

\begin{figure}[h]
\begin{center}
\includegraphics[width=0.45\textwidth]{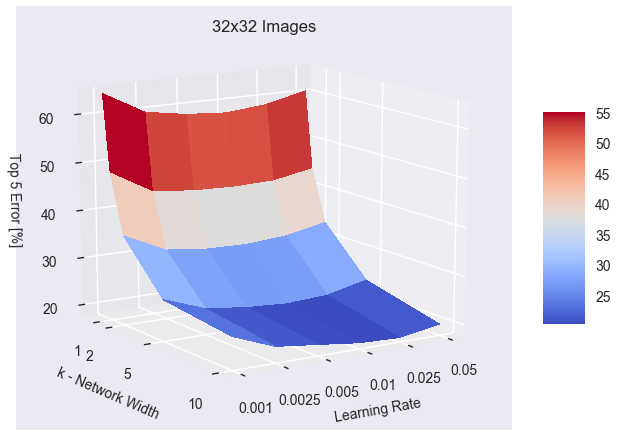}
\includegraphics[width=0.45\textwidth]{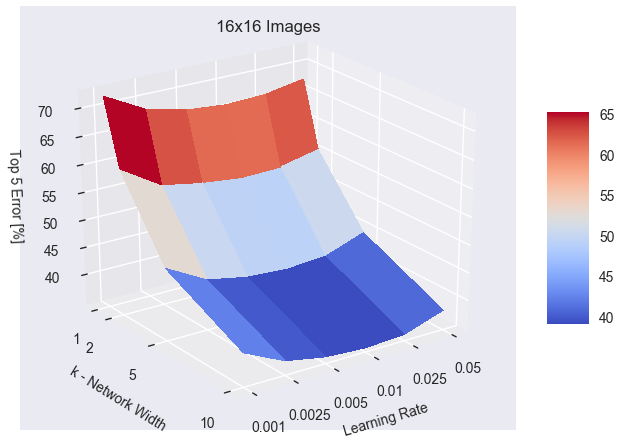}
\end{center}
\caption{The mean Top-5 errors for 32x32 images (left) and 16x16 images (right), as a function of network width $k$ and learning rate.\label{fig:three_d_plots}}
\end{figure}

\begin{figure}[tb]
\begin{center}
\includegraphics[width=0.99\textwidth]{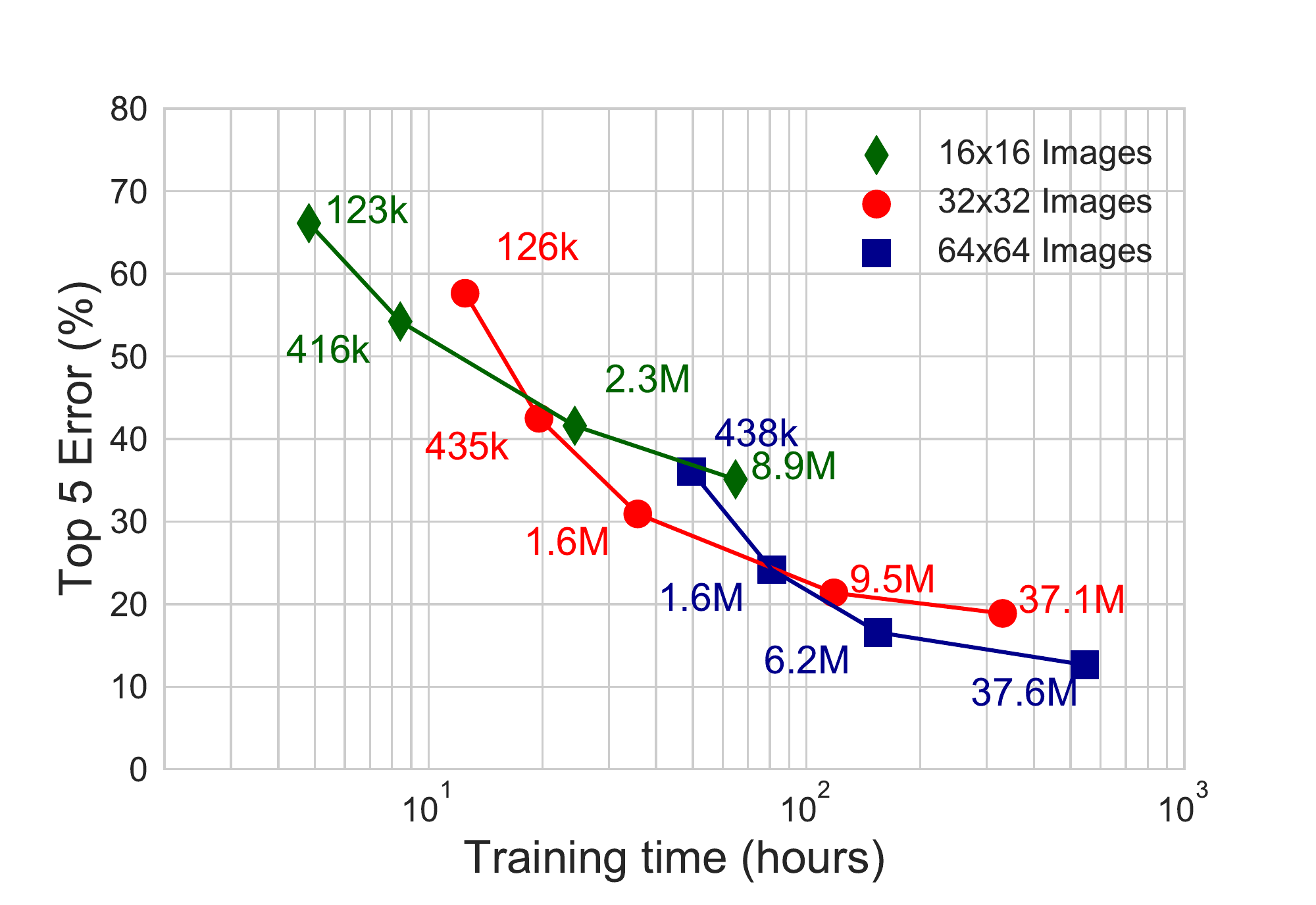}
\includegraphics[width=0.99\textwidth]{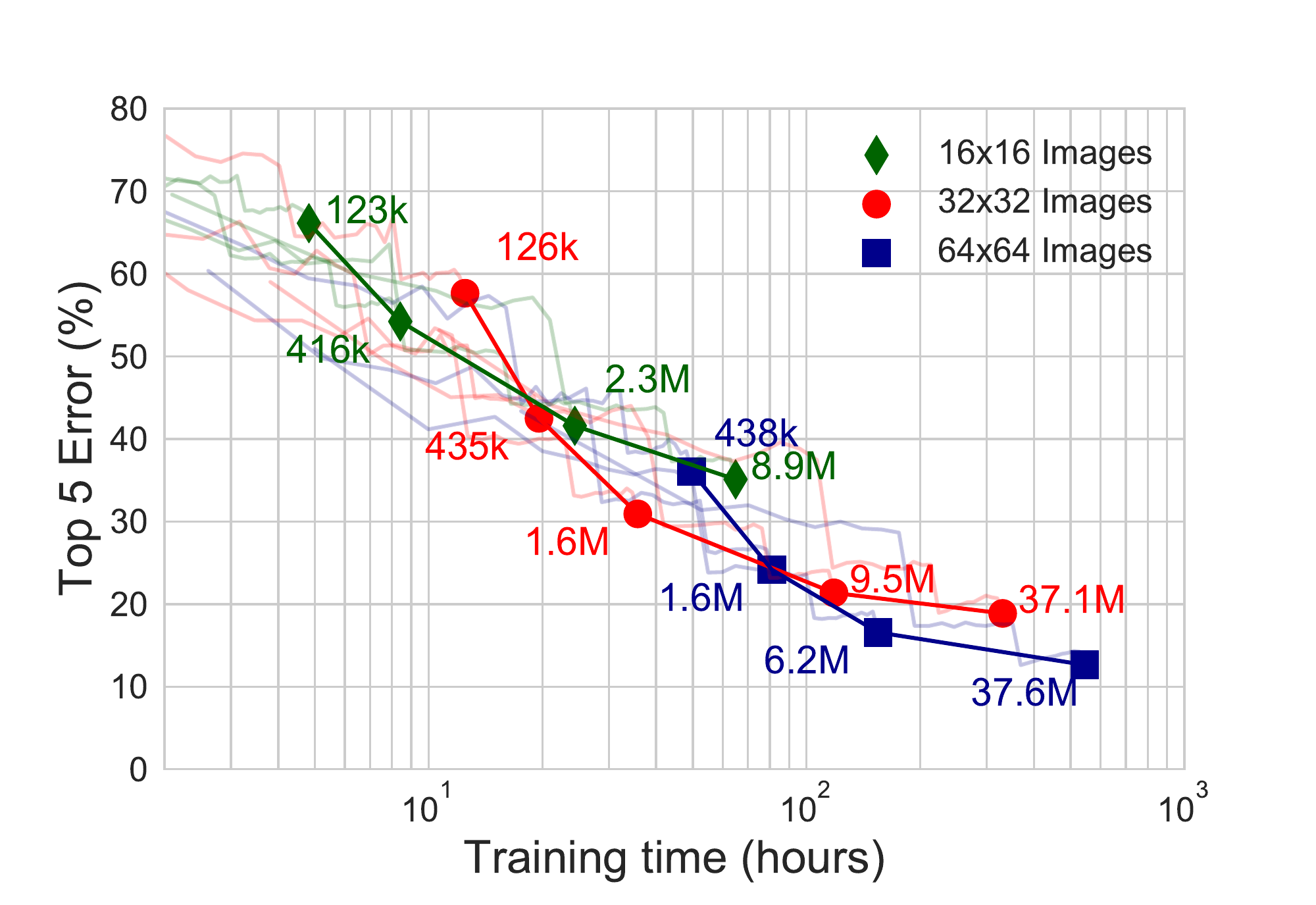}
\end{center}
\caption{The mean Top-5 test error rates according to Table \ref{Table1} for different models (with different number of parameters) vs. training time on a single Titan X GPU. The bottom figure replicates the top one, but also shows semi-transparent curves in the background to represent convergence curves.}
\label{Figure4}
\end{figure}

\item We also investigated the tradeoffs of performance vs.\ training time resulting from different downsampling and network sizes. Figure \ref{Figure4} and Table \ref{Table1} show that both mechanisms for reducing the computational cost should be considered simultaneously to achieve optimal anytime performance. 
An additional mechanism could be to perform warm restarts~\citep{SGDR}, which was shown to substantially improve anytime performance over reductions of the learning rate at regular intervals. %demonstrated that Stochastic Gradient Descent with Warm Restarts can be used to achieve great anytime performance on our proposed ImageNet32x32. \newline
Since the relative ranking of learning rates was consistent across different downsampling and network sizes, we also envision that architecture and hyperparameter search methods could exploit cheap proxies of computationally more expensive setups based on varying these degrees of freedom. Possible methods for exploiting these include~\cite{li2016hyperband,klein2016fast}.

\end{itemize}

\afterpage{\clearpage}
\section{Discussion and Conclusion}

\begin{figure}[tbp]
\begin{center}
\includegraphics[width=0.49\textwidth]{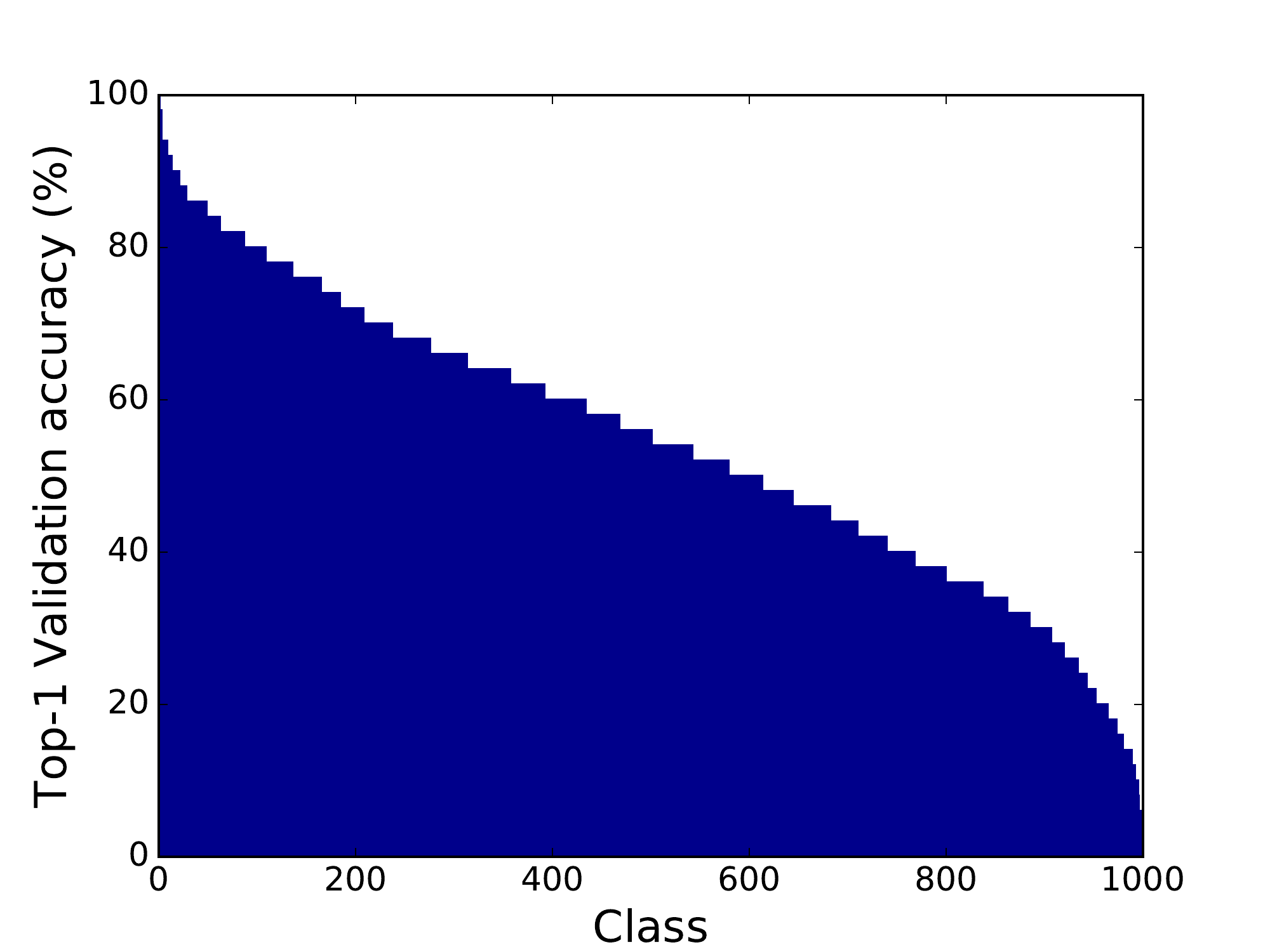}
\includegraphics[width=0.49\textwidth]{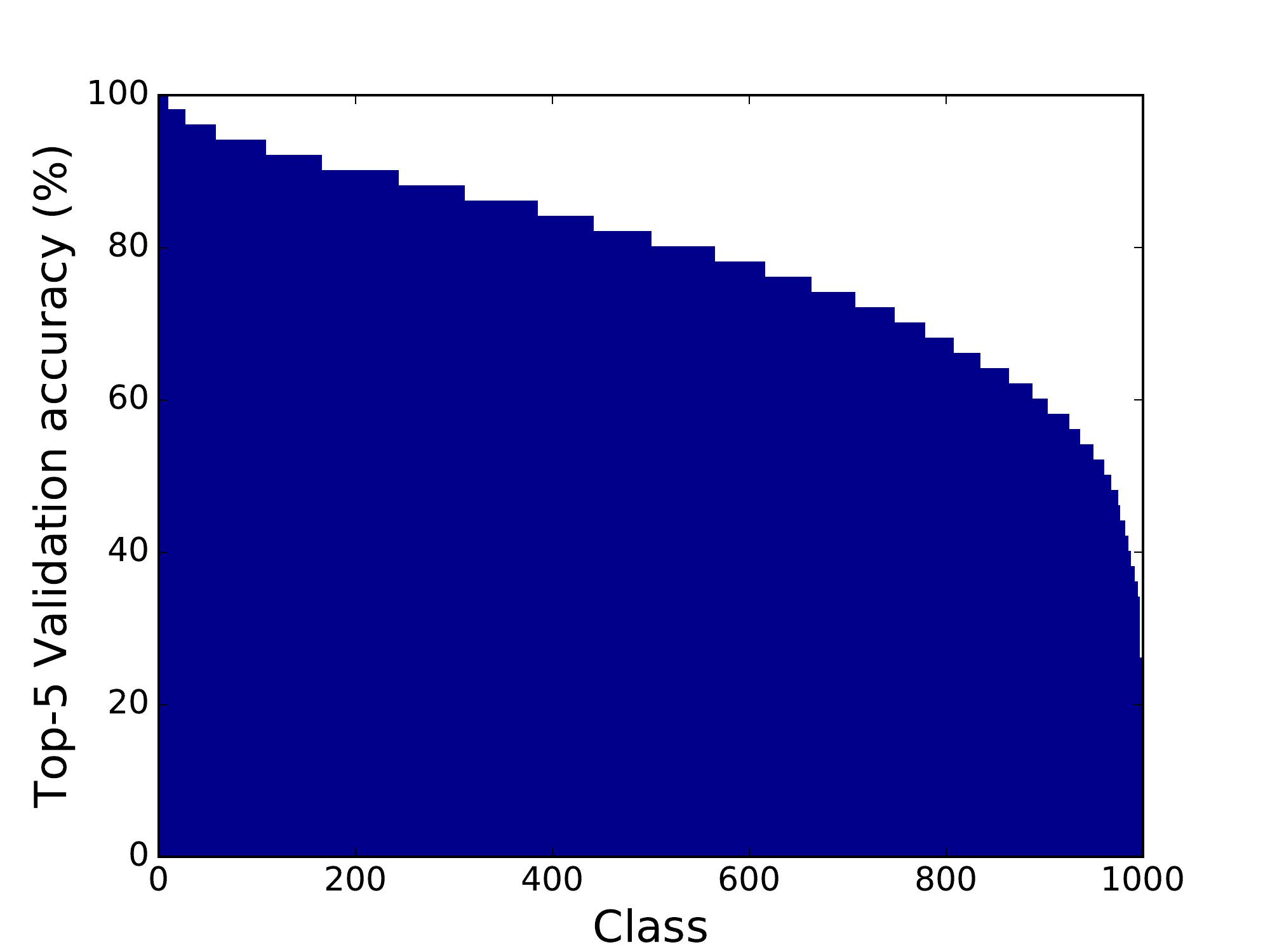}
\end{center}
\caption{Percentage of correct Top-1 (Left) and Top-5 (Right) predictions for different classes obtained by WRN-28-5 on ImageNet32x32. Classes are ordered by this value for better visualization.}
\label{FigureOrd}
\end{figure}

\begin{figure}[tb]
\begin{center}
\includegraphics[width=0.9\textwidth]{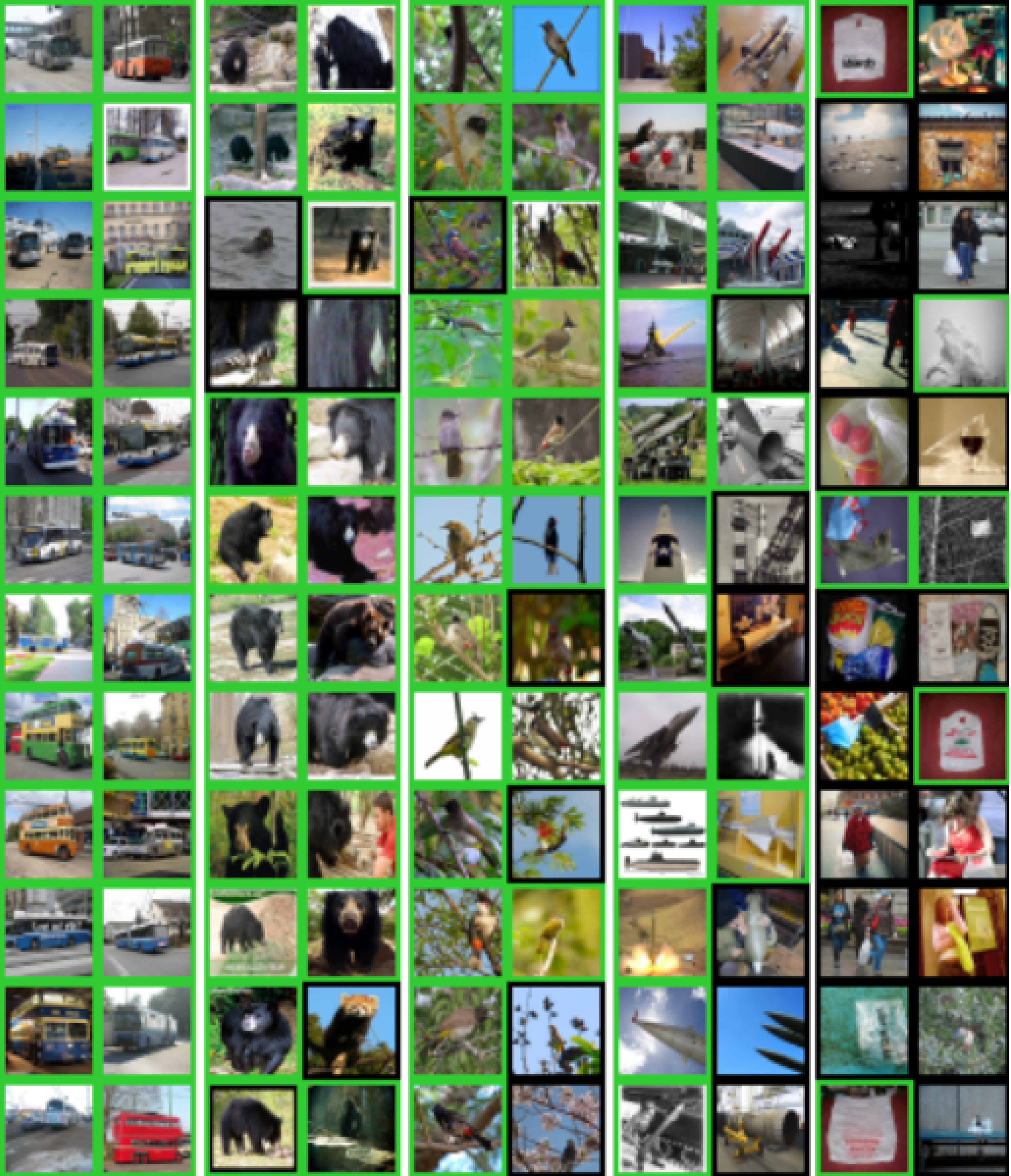}
\end{center}
 \caption{Subset of Imagenet32x32 validation images from classes with 1th (trolleybus, 100\% accuracy), 250th (sloth bear, 88\% accuracy), 500th (bulbul, 80\% accuracy), 750th (projectile, 70\% accuracy) and 1000th (plastic bag, 26\% accuracy) best Top-5 accuracy as given in Figure \ref{FigureOrd}. Green image borders indicate correct Top-5 predictions. The results were obtained by WRN-28-5 on ImageNet32x32.}
\label{FigureOrdExamples}
\end{figure}

Our proposed downsampled versions of the original ImageNet dataset might represent a viable alternative to the CIFAR datasets while dealing with more complex data and classes. Quite surprisingly, even by greatly reducing the resolution of images to 32 $\times$ 32 pixels, one can predict image labels quite well (see also Figure \ref{FigureOrd} and Figure \ref{FigureOrdExamples}). 

Classification of low resolution images might also be of interest when (i) data storage is important (the original ImageNet dataset is 145GB), (ii) the input images are corrupted by noise, or (iii) a small subpart of a high resolution image must be classified. 

We hope that the provided datasets will fill the gap between the CIFAR datasets and the full ImageNet dataset, representing a good benchmark for experimental studies, such as algorithm design, neural network architecture search and hyperparameter optimization. Our preliminary experiments support the hypothesis that findings obtained on smaller networks for lower resolution images may transfer to larger networks for higher resolution images, while being up to 100 times cheaper to obtain. This could be exploited by multi-fidelity methods for architecture and hyperparameter search~\citep{li2016hyperband,klein2016fast}.

\section{Acknowledgement}
This work has partly been supported by the European Research Council (ERC) under the European Union’s Horizon 2020 research and innovation programme under grant no.\ 716721 and by the German Research Foundation (DFG), under the BrainLinksBrainTools Cluster of Excellence (grant number EXC 1086).
The authors acknowledge support by the High Performance and Cloud Computing Group at the Zentrum f\"{u}r Datenverarbeitung of the University of T\"{u}bingen, the state of Baden-W\"{u}rttemberg through bwHPC
and the German Research Foundation (DFG) through grant no INST 37/935-1 FUGG.

%\afterpage{\clearpage}
%\clearpage

\renewcommand\refname{Bibliography}

\bibliographystyle{plainnat}
\bibliography{iclr2017_conference}
\bibliographystyle{iclr2017_conference}

\end{document}